\documentclass[a4paper,10pt]{article}
\usepackage[T1]{fontenc}
\usepackage{lmodern}
\usepackage{amsmath}
\usepackage{graphicx}
\usepackage[margin=25mm]{geometry}
\usepackage[hidelinks]{hyperref}
\hypersetup{
  pdftitle={Stochastic complexity of vectors containing cluster structure},
  pdfauthor={Daniel Nicorici, Olli Yli-Harja, and Jaakko Astola},
  pdfsubject={Originally published in the Proceedings of NSIP 2007},
  pdfkeywords={minimum description length, normalized maximum likelihood,
    stochastic complexity, MDL clustering}
}

\title{Stochastic complexity of vectors containing cluster structure}
\author{Daniel Nicorici\textsuperscript{1} \quad
Olli Yli-Harja\textsuperscript{2} \quad
Jaakko Astola\textsuperscript{2}\\[4pt]
\small \textsuperscript{1}Medicel Oy, Keilaranta 12, FIN-02150 Espoo, Finland\\
\small E-mail: \texttt{Daniel.Nicorici@medicel.com}\\
\small \textsuperscript{2}Institute of Signal Processing, Tampere University of Technology\\
\small P.O. Box 553, FIN-33101 Tampere, Finland\\
\small E-mails: \texttt{Olli.Yli-Harja@tut.fi}, \texttt{Jaakko.Astola@tut.fi}}
\date{\small Originally published in the \emph{Proceedings of the International
Workshop on Nonlinear Signal and Image Processing}\\
\emph{(NSIP 2007)}, Bucharest, Romania, 10--12 September 2007, pp.~164--169.}

\begin{document}
\maketitle
\begin{abstract}
This paper studies the problem of computing the stochastic
probability (shortest code length) of the encoded vectors containing
cluster structure using Normalized Maximum Likelihood (NML) model.
This is of great theoretical and practical importance in data
clustering based on Minimum Description Length (MDL) principle, such
as for estimating the best number of clusters and best cluster
structure for the data. Straightforward computation of the shortest
code length of the vector containing cluster structure based on the
NML model requires polynomial time with respect to the size of the
vector and number of clusters. We show that this is a tractable
problem by introducing a recursion formula for the efficient
computation of normalizing constant from the NML model. The time
complexity of the new formula is linear opposed to previous
polynomial time with respect to the size of the vector and number of
clusters.

\medskip
\noindent\textbf{Keywords:} minimum description length, normalized maximum
likelihood, stochastic complexity, MDL clustering, estimating number
of clusters, finding cluster structure.
\end{abstract}

\section{Introduction}
\label{sec:Introduction}

The MDL principle for statistical model selection and statistical
inference is based on the simple idea that the best way to capture
regular features in data is to construct a model in a certain class
which permits the shortest description of the data and the model
itself~\cite{Rissanen1978,Rissanen1983,Rissanen2000,Rissanen2001,Rissanen2005}.
This principle chooses a model such that it trades off
between goodness-of-fit on the observed data with
``complexity'' of the model~\cite{Grunwald2005-Tutorial}. Also, the
MDL approach fits models to the data and no assumption that the data
are a sample from a ``true'' random variable is
needed~\cite{Rissanen2000,Rissanen2005}. This eliminates the
difficulty in other approaches to modeling that the more complex
model is fitted to the data the better estimate of the data one
gets~\cite{Rissanen2000,Rissanen2001}. The MDL
principle~\cite{Rissanen1978,Rissanen2000,Rissanen2005} has been
widely used in statistical inference in an asymptotically justified
form~\cite{Tabus2003-Classification}.

Recently, a more efficient form of MDL principle based on the
normalized maximum likelihood model, which effects a universal
sufficient statistics decomposition to separate noise from the
learnable information in data has been
found~\cite{Tabus2003-Classification}. The NML model provides a
better yardstick for comparison of performance of different model
classes than other alternative ways of computing the code length,
e.g. two-part
codes~\cite{Rissanen2000,Rissanen2001,Tabus2003-Classification}. The
stochastic complexity, based on the NML model, is the shortest
description length of a given data for a given model
class~\cite{Kontkanen2005-Tree}.

The scenario we follow is that the encoder and the decoder agree
beforehand on the NML model for a specific class of models that is
being used to design the codes for encoding the message, e.g. the
vector containing the cluster structure. The optimal length of the
message is well known in information theory and it can be achieved
based on practical coding methods~\cite{Tabus2003-Classification}.
However, our goal is to compute the code length of the encoded
message and not to write down the encoded message. We choose the
model which gives the smallest encoded message based on the MDL
principle.

Clustering is one of the central concepts in the field of
unsupervised data analysis. The problem of finding the cluster
structure and number of clusters for the given data, based on the
MDL principle, when no information other than the observed values is
available is known as MDL clustering and its goal is to partition
the data into several non-hierarchical groups of
items~\cite{Kontkanen2005}. The approach of the MDL clustering is
based on the idea that a good clustering is such that one can encode
the cluster structure together with the data based on the NML models
so that the resulting total code length is
minimized~\cite{Kontkanen2005}. The models, representing the
different ways of clustering the data, are compared based on the
code lengths. The problems of finding the number of clusters and the
cluster structure are solved simultaneously by choosing the best
model, which gives the shortest code length for the data.

The efficient encoding based on NML model of the vector containing
the cluster structure, also called clustering vector throughout this
study, is of great importance in the MDL clustering framework. Our
previous study~\cite{Nicorici2005-SSP} introduced an
efficient NML model for encoding the clustering vector and a
polynomial time method to compute its NML code length. This had been
used in our previous
work~\cite{Nicorici2005-SSP,Nicorici2006-GENSIPS} for finding the
number of clusters and the cluster structure in gene expression data
from microarray experiments. The previously introduced polynomial
time method, which is similar to Kontkanen {\it et
al.}~\cite{Kontkanen2005}, is infeasible for large of even moderate
size data sets. In this study, our main goal is to derive a linear
time recursive method for the computation of the NML code length of
the encoded clustering vector, based on properties of generating
functions~\cite{Szpankowski2001book}. The properties of the
generating functions have been previously used for finding
asymptotic expansions for sums arising in coding
theory~\cite{Szpankowski1995asymptotics,Szpankowski1998asymptotics,Szpankowski2001book}
and for computing the stochastic complexity in the case of
multinomial data~\cite{Kontkanen2005-Tree}.

In section 2 we introduce introduce the notation and review MDL
clustering. The NML model for encoding clustering vector is
presented in section 3. The properties of generating function is
presented in section 4 and the new recursion formula for computing
the NML code length of the encoded clustering vector is derived
using a generating function in section 5. Finally, section 6 gives
the concluding remarks.


\section{MDL clustering}
\label{sec:MLDclustering}

Here we present the MDL clustering approach of Kontkanen {\it et
al.}~\cite{Kontkanen2005} for data clustering. Let us consider a
data set $\mathbf{x}^n=(\mathbf{x}_1,\ldots,\mathbf{x}_n)$
consisting of $n$ column vectors, where
$\mathbf{x}_i=(x_{i1},\ldots,x_{iq})^T$ and $i=1,\ldots,n$.
Clustering of the data set $\mathbf{x}^n$ is defined as a
partitioning of the data into mutually exclusive subsets, the union
of which forms the data set. We denote a clustering by using the
clustering vector $\mathbf{y}^n=(y_1,\ldots,y_n)$ where
$y_i=\{1,\ldots,m\}$ and $y_i=k$ where $k \in \{1,\ldots,m\}$ if and
only if $\mathbf{x}_i$ belongs to the cluster $k$. The number of
clusters is denoted by $m$, where $m=1,\ldots,n$.

The MDL code length (two
part-code)~\cite{Rissanen1978,Kontkanen2005} of the encoded data
$\mathbf{x}^n$ together with the clustering vector $\mathbf{y}^n$
considering the model $\mathcal{M}_m$~\cite{Kontkanen2005} is
\begin{equation} \label{eq:MDL2}
\mathcal{L}(\mathbf{x}^n,\mathbf{y}^n|\mathcal{M}_m) =
\mathcal{L}(\mathbf{y}^n|\mathcal{M}_m) +
\mathcal{L}(\mathbf{x}^n|\mathbf{y}^n),
\end{equation} where the first term gives the cost of encoding
the clustering vector $\mathbf{y}^n$ and the last term gives the
cost of encoding $\mathbf{x}^n$. In the case that the data
$\mathbf{x}^n$ contains discrete values one can compute its code
length $\mathcal{L}(\mathbf{x}^n|\mathbf{y}^n)$ as
in~\cite{Kontkanen2005,Nicorici2005-GENSIPS}. When the data contains
no quantized values the approach in~\cite{Nicorici2006-GENSIPS} can
be used for computing $\mathcal{L}(\mathbf{x}^n|\mathbf{y}^n)$.

\section{NML model for encoding clustering vector}

In the approach of Kontkanen {\it et al.} \cite{Kontkanen2005} the
clustering vector $\mathbf{y}^n$ is encoded considering all possible
$m$-ary sequences of length $n$. We introduced a refinement in the
NML model for computing code length of the encoded clustering vector
$\mathbf{y}^n$ by taking into account more faithfully the number of
clustering vectors. This allows us to encode more efficiently the
clustering vector using the NML model and to discriminate better
between different models.

The NML approach is used for encoding the clustering vector
$\mathbf{y}^n$ by postulating a simple parametric model
$P(\mathbf{y}^n;\Theta(\mathbf{y}^n))$. The clustering vector
$\mathbf{y}^n$ contains in general $h_i$ values of $i$ and for
$\mathbf{y}^n$ the ML estimate of
$\Theta=\{\theta_1,\ldots,\theta_q\}$ is
$\hat{\Theta}(\mathbf{y}^n)=\{\hat{\theta}_1(\mathbf{y}^n),\ldots,\hat{\theta}_q(\mathbf{y}^n)\}$,
where $\hat{\theta}_i(\mathbf{y}^n)=\frac{h_i}{n}$. The probability
of the $\mathbf{y}^n$ becomes
$P(\mathbf{y}^n;\hat{\Theta}(\mathbf{y}^n)) = \prod_{i=1}^{m}
{(\frac{h_i}{n})}^{h_i}$. The NML model is known to be a solution of
two minmax problems, which gives it strong optimality properties
\cite{Rissanen2001}. The normalized maximum likelihood is
\begin{equation}\label{eq:probNML}
\hat{P}(\mathbf{y}^n;\hat{\Theta}(\mathbf{y}^n)) =
\frac{\prod^{m}_{i=1} {\left(\frac{h_i}{n}\right)}^{h_i}} {C_n(m)},
\end{equation}
where
\begin{equation} \label{eq:space}
C_n(m) = \sum_{\mathbf{t}^n \in \mathcal{F}^n_m}
P(\mathbf{t}^n;\hat{\Theta}(\mathbf{t}^n)),
\end{equation}
is the normalizing constant for the probability
$P(\mathbf{y}^n;\hat{\Theta}(\mathbf{y}^n))$ and $\mathcal{F}^n_m$
is chosen to include all the possible sequences representing
clustering vectors $\mathbf{t}^n$ containing exactly $m$ clusters
and that are unique among them. Two clustering vectors are
considered not unique when they describe the same cluster structure,
i.e. RAND index$=1$. The RAND index~\cite{Hubert1985} ranges between
0 and 1 and it is a measure of agreement between alternative data
partitions (data clusters).

For instance, one has for $n=4$ and $m=2$, the following space
$\mathcal{F}^4_2$ of $7$ unique clustering vectors
\begin{eqnarray*}
&&\left(1,2,2,2\right)\\
&&\left(2,1,2,2\right)\\
&&\left(2,2,1,2\right)\\
&&\left(2,2,2,1\right)\\
&&\left(1,1,2,2\right)\\
&&\left(1,2,1,2\right)\\
&&\left(1,2,2,1\right)
\end{eqnarray*}
instead of all $14$ possible clustering vectors
\begin{eqnarray*}
&&\left(1,1,1,2\right) \\
&&\left(1,1,2,1\right) \\
&&\left(1,1,2,2\right) \\
&&\left(1,2,1,1\right) \\
&&\left(1,2,1,2\right) \\
&&\left(1,2,2,1\right) \\
&&\left(1,2,2,2\right) \\
&&\left(2,1,1,1\right) \\
&&\left(2,1,1,2\right) \\
&&\left(2,1,2,1\right) \\
&&\left(2,1,2,2\right) \\
&&\left(2,2,1,1\right) \\
&&\left(2,2,1,2\right) \\
&&\left(2,2,2,1\right)
\end{eqnarray*} which are not all unique
between them, e.g. the clustering vectors $\left(1,1,1,2\right)$ and
$\left(2,2,2,1\right)$ represent the same way of clustering the data
(the first three samples are clustered together and the fourth
sample is clustered separately), i.e. RAND Index$=1$.

One can notice that the normalization constant in (\ref{eq:space})
for encoding the clustering vector $\mathbf{y}^n$ with $m$ unique
clusters is~\cite{Nicorici2005-SSP},
\begin{equation} \label{eq:NormalizationDaniel}
C_n(m) = \frac{C_n^*(m)}{m!},
\end{equation}
where $C_n^*(m)$ is the normalizing constant for the NML model for
encoding $m$-ary sequences of length $n$ such that all symbols from
the $m$-alphabet appear at least once in every sequence; $C_n^*(m)$
includes all possible such $m$-ary sequences. Thus one has
\begin{equation}\label{eq:NormalizationFactor}
C_n^*(m)=\sum_{\substack{h_1 + \ldots + h_m = n\\h_1, \ldots, h_m
\geq 1}} \frac{n!}{h_1! \ldots h_m!} \prod^{m}_{i=1}{\left(
\frac{h_i}{n} \right)}^{h_i}.
\end{equation}
An efficient way of computing the
$C_n^*(m)$~\cite{Nicorici2005-SSP,Szpankowski2001book,Kontkanen2005}
is as following
\begin{equation}\label{eq:Recursive}
C_n^*(m) = \sum_{i=1}^{n-1} \frac{n!}{i!(n-i)!}
{\left(\frac{i}{n}\right)}^{i} {\left(\frac{n-i}{n}\right)}^{n-i}
C_i^*(m-1),
\end{equation}
where $C_0^*(m)=1$ and $C_n^*(1)=1$. The values for $C_n^*(m)$  and
$C_n^*(m)$ can be pre-computed and tabulated to speed up the
computations.

The code length, in bits, of the encoded clustering vector
$\mathbf{y}^n$ with $m$ clusters based on the NML model is
\begin{eqnarray}\label{eq:codelength}
\mathcal{L}(\mathbf{y}^n|\mathcal{M}_m) &=& -\log_2
\hat{P}(\mathbf{y}^n;\hat{\Theta}(\mathbf{y}^n))\nonumber\\
&=& \log_2 C_n(m)-\log_2\prod^m_{i=1} \left( \frac{h_i}{n}
\right)^{h_i}.
\end{eqnarray}
For instance, for a clustering vector $\mathbf{y}^n$ with $n=m$
clusters (each sample is clustered separately), one has from
(\ref{eq:codelength}) that
$\mathcal{L}(\mathbf{y}^n|\mathcal{M}_n)=0$, which is what one would
expect. In this case no additional information is needed for
encoding the clustering vector beside the order of the model. It is
enough for the decoder to know that $n=m$ in order to decode the
clustering vector. When the method from~\cite{Kontkanen2005} for
$n=m$ is used, one has $\mathcal{L}(\mathbf{y}^n|\mathcal{M}_n)>0$
and the code length of the encoded clustering vector is longer and
not so efficient as in~(\ref{eq:codelength}).

\section{Generating Function}

The generating functions are a popular analytic tool in establishing
recurrence formulas, finding asymptotic expansions, and proving
combinatorial identities~\cite{Szpankowski2001book}.

The generating function of a sequence $\{a_n\}=(a_0,a_1,\ldots)$ is
defined~\cite{Szpankowski2001book} as
\begin{equation}
A(z)=\sum_{n=0}^\infty a_n z^n.
\end{equation}
An useful tree function that generates the sequence
$\{\frac{n^{n-1}}{n!}\}$ is Cayley's tree function
$T(z)$~\cite{Szpankowski2001book,Szpankowski1995asymptotics,Szpankowski1998asymptotics}
where
\begin{equation}\label{eq:TzSum} T(z)=\sum_{n=1}^\infty
\frac{n^{n-1}}{n!}z^{n},
\end{equation} and one of its basic properties is
\begin{equation}\label{eq:TzE}
T(z)=z\cdot e^{T(z)}.
\end{equation}

Differentiating (\ref{eq:TzSum}) and multiplying with $z$ yields
\begin{equation}\label{eq:TzSumDiff}
z\cdot T^\prime (z)=\sum_{n=0}^\infty
\frac{n^{n}}{n!}z^{n}-1=B(z)-1,
\end{equation} where
\begin{equation}\label{eq:Bz}
B(z)=\sum_{n=0}^\infty \frac{n^{n}}{n!}z^{n},
\end{equation} generates the sequence $\{\frac{n^n}{n!}\}$.
From (\ref{eq:TzSumDiff}) one has
\begin{equation}\label{eq:BzTzDiff}
B(z)=z T^\prime(z)+1.
\end{equation}

Also, by differentiating (\ref{eq:TzE}) one has
\begin{equation}\label{eq:TzEDiff}
T^\prime(z)=e^{T(z)}+z\cdot e^{T(z)}\cdot T^\prime(z).
\end{equation}
By plugging (\ref{eq:TzE}) in (\ref{eq:TzEDiff}) one has
\begin{equation}\label{eq:zTzDiff}
z\cdot T^\prime(z)=\frac{T(z)}{1-T(z)}.
\end{equation}
Also, by plugging (\ref{eq:zTzDiff}) in (\ref{eq:TzSumDiff}) one has
\begin{equation}\label{eq:BzFinal}
B(z)=\frac{1}{1-T(z)},
\end{equation}and
\begin{equation}\label{eq:BzStarFinal}
B(z)-1=\frac{T(z)}{1-T(z)}.
\end{equation}

To show the connection between $B(z)$ and $C_n^*(m)$ we compute
$(B(z)-1)^2$, which yields
\begin{eqnarray} \label{eq:B2m1}
&&\left(B(z)-1\right)^2= \sum_{h_1=1}^\infty \frac{h_1^{h_1}}{h_1!}z^{h_1}\sum_{h_2=1}^\infty \frac{h_2^{h_2}}{h_2!}z^{h_2}\nonumber\\
&&=\sum_{n=2}^\infty \left(\sum_{\substack{h_1+h_2=n\\h_1,h_2 \geq
1}}
\frac{h_1^{h_1}}{h_1!}\frac{h_2^{h_2}}{h_2!}\right)z^{h_1+h_2}\nonumber\\
&&=\sum_{n=2}^\infty \frac{n^n}{n!}
\left(\sum_{\substack{h_1+h_2=n\\h_1,h_2 \geq 1}}
\frac{n!}{h_1!h_2!} \left(\frac{h_1}{n}\right)^{h_1} \left(\frac{h_2}{n}\right)^{h_2}\right)z^n\nonumber\\
&&=\sum_{n=2}^\infty \frac{n^n}{n!} \left(\sum_{h_1=1}^{n-1}
\binom{n}{h_1} \left(\frac{h_1}{n}\right)^{h_1} \left(\frac{n-h_1}{n}\right)^{n-h_1}\right)z^n\nonumber\\
&&=\sum_{n=2}^\infty \frac{n^n}{n!} C_n^*(2) z^n,
\end{eqnarray} where the last equality follows
from~(\ref{eq:Recursive}) for $m=2$. Therefore it is straightforward
to prove this to
\begin{equation} \label{eq:Bmm1}
\left(B(z)-1\right)^m=\sum_{n=2}^\infty \frac{n^n}{n!} C_n^*(m) z^n,
\end{equation} which generates the sequence
$\{\frac{n^n}{n!}C_n^*(m)\}$. Thus our generating function of
interest is $(B(z)-1)^m$.

Plugging~(\ref{eq:BzStarFinal}) in~(\ref{eq:Bmm1}) yields
\begin{equation}\label{eq:BzStarPowerM}
(B(z)-1)^m=\frac{T^m(z)}{(1-T(z))^m}=\sum_{n=2}^{\infty}\frac{n^n}{n!}C_n^*(m)z^n.
\end{equation}

\section{Recursion formula}

Based on~(\ref{eq:BzStarPowerM}), let define a special tree function
\begin{equation}\label{eq:Tree}
\frac{T^{m-1}(z)}{(1-T(z))^{m-1}}=\sum_{n=1}^{\infty}C^*_n(m-1)\frac{n^n}{n!}z^n.
\end{equation}

Differentiating the right-hand term of~(\ref{eq:Tree}) and
multiplying with $z$, one has
\begin{equation}\label{eq:TreeFirst}
z\cdot\frac{d}{dz}\sum_{n=1}^{\infty}C_n^*(m-1)\frac{n^n}{n!}z^n=\sum_{n=2}^{\infty}n
C_n^*(m-1)\frac{n^n}{n!}z^{n}.
\end{equation}

\begin{figure}[htb]
  \centering
  \includegraphics[width=0.7\linewidth]{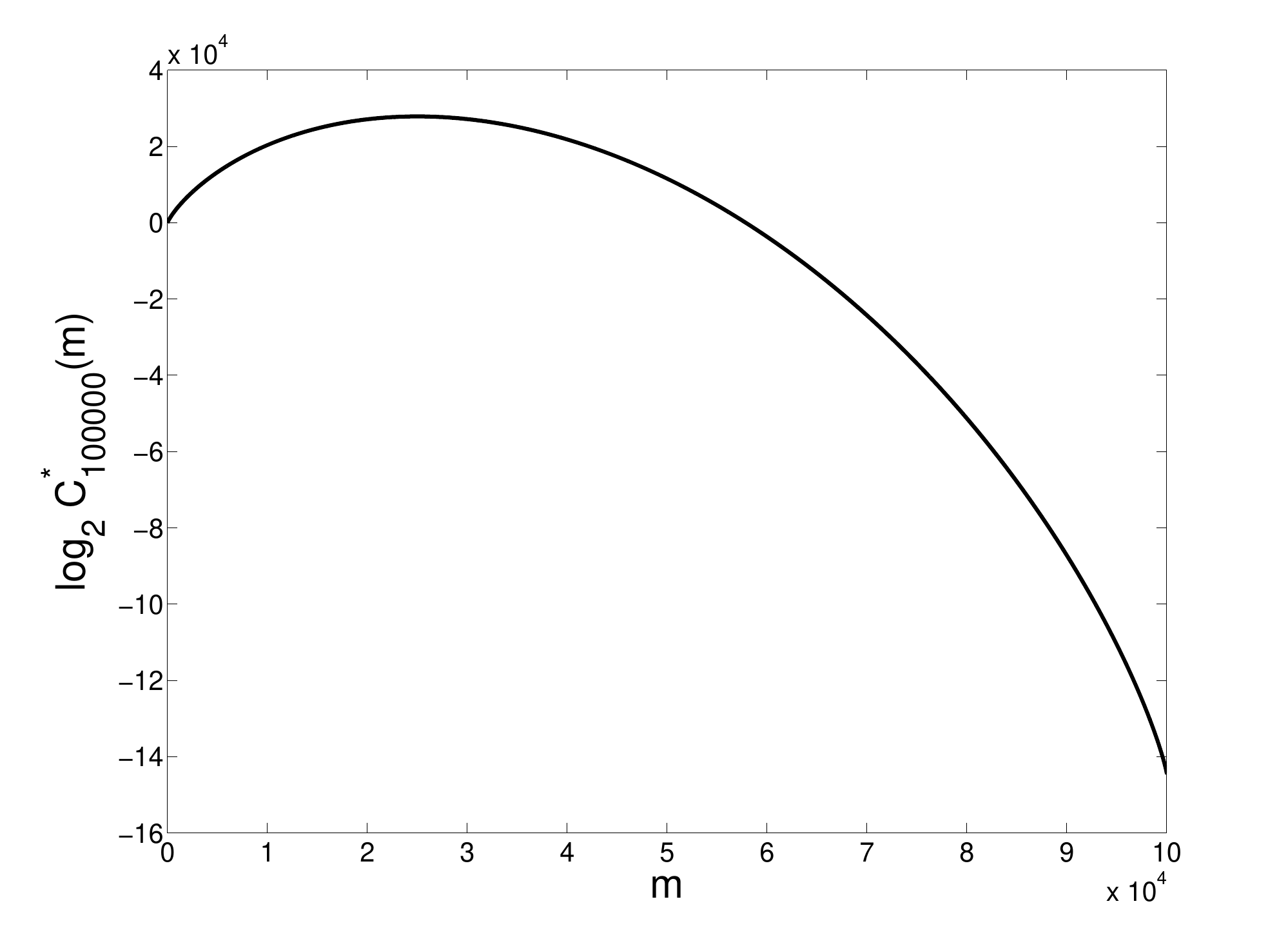}
\caption{The NML normalizing constant $C_n^*(m)$ that includes all
possible $m$-ary sequences of length $n$ and all symbols from the
$m$-alphabet appear at least once in every sequence, where
$n=100000$ and $m=1\ldots n$.} \label{fig:CStarNM}
\end{figure}

Also, differentiating the left-hand term of~(\ref{eq:Tree}) and
multiplying with $z$, one has
\begin{eqnarray}\label{eq:TreeSecond}
&&z\cdot\frac{d}{dz}\frac{T^{m-1}(z)}{(1-T(z))^{m-1}}=\frac{T^m(z)}{(1-T(z))^m}\frac{m-1}{T^2(z)}zT^\prime(z)\nonumber\\
&&=(m-1)\frac{T^m(z)}{(1-T(z))^m}\frac{1}{\left(1-T(z)\right)T(z)}\nonumber\\
&&=(m-1)\frac{T^{m+1}(z)}{(1-T(z))^{m+1}}+2(m-1)\frac{T^m(z)}{(1-T(z))^m}\nonumber\\
&&+(m-1)\frac{T^{m-1}(z)}{(1-T(z))^{m-1}}\nonumber\\
&&=(m-1)\sum_{n=2}^{\infty}\frac{n^n}{n!}C_n^*(m+1)z^n\nonumber\\
&&+2(m-1)\sum_{n=2}^{\infty}\frac{n^n}{n!}C_n^*(m)z^n\nonumber\\
&&+(m-1)\sum_{n=2}^{\infty}\frac{n^n}{n!}C_n^*(m-1)z^n.
\end{eqnarray}
Comparing the coefficients of $z^n$ in~(\ref{eq:TreeFirst})
and~(\ref{eq:TreeSecond}), one has
\begin{eqnarray} \label{eq:TreeFinal}
n C_n^*(m-1)&=&(m-1) C_n^*(m+1)+2(m-1)C_n^*(m)\nonumber\\
&&+(m-1)C_n^*(m-1),
\end{eqnarray}and it yields
\begin{equation}\label{eq:CnmStarRecurrenceFinal}
C_n^*(m)=\frac{m}{n-m}\left(C_n^*(m+2)+2C_n^*(m+1)\right),
\end{equation}which holds for $1<m<n-1$. For recursively
computing~(\ref{eq:CnmStarRecurrenceFinal}) one needs to know
$C_n^*(n)$ and $C_n^*(n-1)$ which are straightforward to compute
using~(\ref{eq:NormalizationFactor}). Therefore one has
\begin{equation}\label{eq:CnnStar}
C_n^*(n)=\frac{n!}{n^n},
\end{equation}
and
\begin{equation}\label{eq:Cnnm1Star}
C_n^*(n-1)=2(n-1)\frac{n!}{n^n}.
\end{equation}
The computation of $C_n^*(m)$ starts with $C_n^*(n)$ and
$C_n^*(n-1)$ and then \mbox{$C_n^*(n-2)$},\ldots,$C_n^*(m)$ are computed
using the recursion formula~(\ref{eq:CnmStarRecurrenceFinal}). The
plot of the normalizing constant $C_n^*(m)$ computed using the
recursion formula~(\ref{eq:CnmStarRecurrenceFinal}) is represented
in Figure~\ref{fig:CStarNM}. It can be seen that the maximum of
$C_n^*(m)$ is achieved for $\lfloor \frac{n}{4} \rfloor+1$ for a
given $n$.

From~(\ref{eq:NormalizationDaniel}),~(\ref{eq:CnnStar}),~(\ref{eq:Cnnm1Star}),
and~(\ref{eq:CnmStarRecurrenceFinal}) one has the recursion formula
for normalizing constant $C_n(m)$ as follows,
\begin{equation}\label{eq:CnmRecurrenceFinal}
C_n(m)=\frac{m(m+1)}{n-m}\left((m+2)C_n(m+2)+2C_n(m+1)\right),
\end{equation}which holds for $1<m<n-1$, where
\begin{equation} \label{eq:Cnn}
C_n(n)=\frac{1}{n^n},
\end{equation}
and
\begin{equation} \label{eq:Cnnm1}
C_n(n-1)=\frac{2n(n-1)}{n^n}.
\end{equation}

\begin{figure}[htb]
  \centering
  \includegraphics[width=0.7\linewidth]{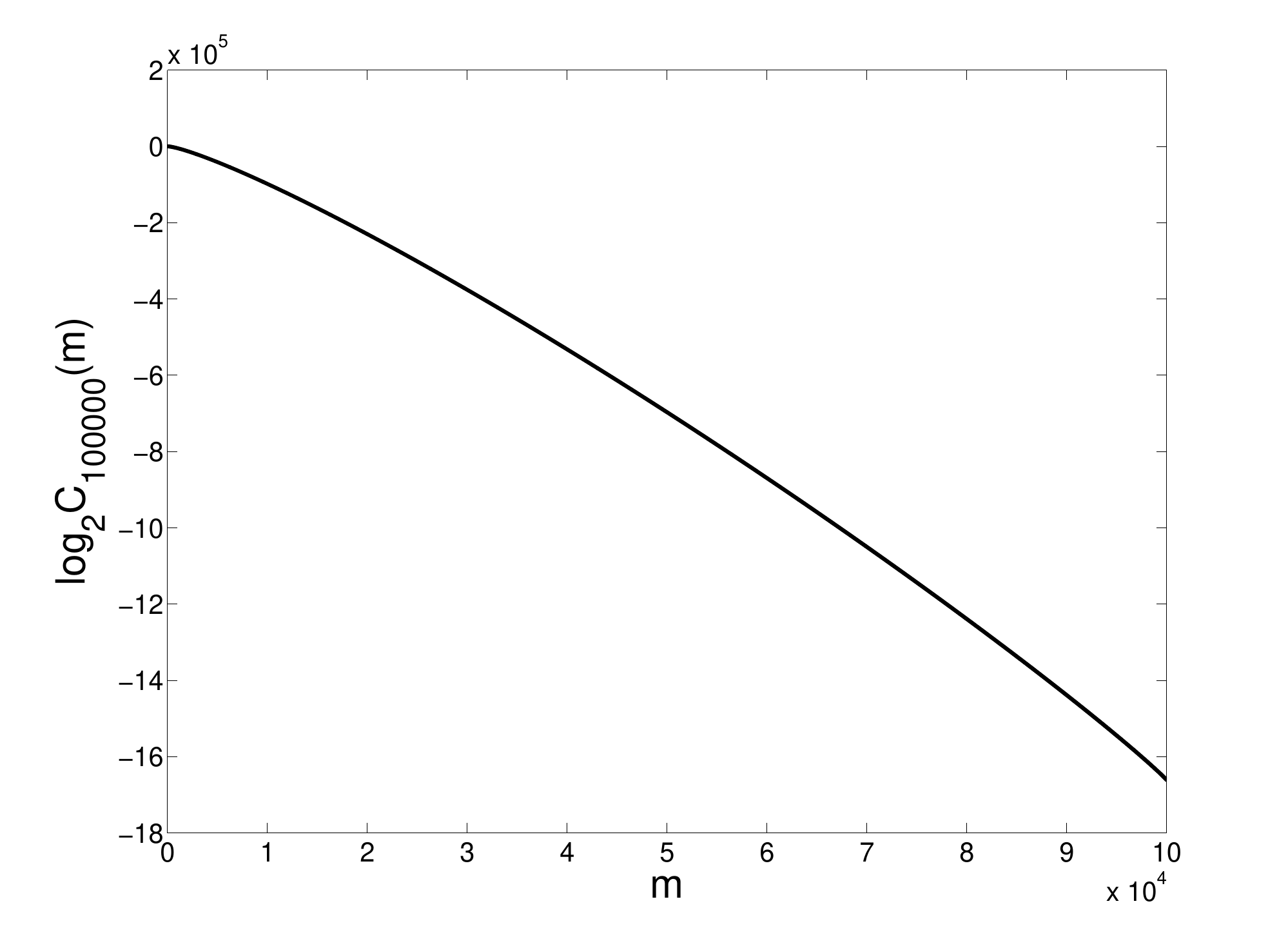}
\caption{The NML normalizing constant $C_n(m)$ that includes all
possible clustering vectors of length $n$ with $m$ unique clusters,
where $n=100000$ and $m=1\ldots n$.} \label{fig:CNM}
\end{figure}

The computation of $C_n(m)$ starts with $C_n(n)$ and
$C_n(n-1)$ using~(\ref{eq:Cnn}) and~(\ref{eq:Cnnm1}), and then
\linebreak
\mbox{$C_n(n-2)$},\ldots,$C_n(m)$ are computed using the recursion
formula~(\ref{eq:CnmRecurrenceFinal}). The plot of the normalizing
constant $C_n(m)$ computed using the recursion
formula~(\ref{eq:CnmRecurrenceFinal}) is represented in
Figure~\ref{fig:CNM}. The whole computation has a linear time
complexity which is a major improvement compared to the previous
method~\cite{Nicorici2005-SSP}.

\section{Concluding remarks}

The efficient encoding based on NML model of the encoded vector
containing the cluster structure and efficient computation of its
NML code length are of great importance in the MDL clustering
framework. In this study we introduced a linear time complexity
method for fast computation of the NML code length of the encoded
vector containing cluster structure. The previous method has a
polynomial time complexity with respect of the size and numbers of
clusters in the vector. The newly introduced method is based on a
recursion formula for efficient computation of normalizing constant
from the NML model for clustering vectors.

\bibliographystyle{IEEEbib}
\bibliography{biblio}

\end{document}